\documentclass[runningheads]{llncs}

\usepackage[T1]{fontenc}
\usepackage{graphicx}

\usepackage{amsmath}
\usepackage{amssymb}
\usepackage{booktabs}
\usepackage{tabularx}
\usepackage{algorithm2e}
\usepackage{xcolor}
\usepackage[hidelinks]{hyperref}

\begin{document}

\title{Heterogeneous Debate Engine:\\ Identity-Grounded Cognitive Architecture for \\ Resilient LLM-Based Ethical Tutoring}

\titlerunning{Heterogeneous Debate Engine}

\author{Jakub Masłowski\orcidID{0009-0005-4597-6335} \and \\
Jarosław A. Chudziak\orcidID{0000-0003-4534-8652}}

\authorrunning{J. Masłowski and J. Chudziak}

\institute{Institute of Computer Science, \\ Warsaw University of Technology, Poland\\
\email{\{jakub.maslowski2.stud, jaroslaw.chudziak\}@pw.edu.pl}}

\maketitle

\begin{abstract}

Large Language Models (LLMs) are being increasingly used as autonomous agents in complex reasoning tasks, opening the niche for dialectical interactions.
However, Multi-Agent systems implemented with systematically unconstrained systems systematically undergo semantic drift and logical deterioration and thus can hardly be used in providing ethical tutoring where a precise answer is required. Current simulation often tends to degenerate into dialectical stagnation, the agents degenerate into recursive concurrence or circular arguments.
A critical challenge remains: how to enforce doctrinal fidelity without suppressing the generative flexibility required for dialectical reasoning?
To address this niche, we contribute the Heterogeneous Debate Engine (HDE), a cognitive architecture that combines Identity-Grounded Retrieval-Augmented Generation (ID-RAG) for doctrinal fidelity and Heuristic Theory of Mind for strategic opponent modeling.
Our evaluation shows that architectural heterogeneity is a crucial variable to stability: contrary doctrinal initializations (e.g., Deontology vs. Utilitarianism) have increased the Argument Complexity Scores of students by an order of magnitude, over baselines. These findings validate the effectiveness of ID-RAG and Heuristic ToM as architectural requirements in maintaining high-fidelity (adversarial) pedagogy.

\end{abstract}

\section{Introduction}
\label{sec:introduction}

The accelerating growth of Artificial Intelligence is leading to a transition to the Agentic World, in which systems based on AI have advanced to no longer be mere tools, but functioning as actors with goals and capable of complex cognition and thought processes \cite{hou2025halohierarchicalautonomouslogicoriented,hayashi2025planning,west2025abductactpredictscaffolding}. This paradigm provides immense opportunities to scaffold critical thinking in the field of education, including the critical thinking \cite{peng2025kele,hou2025eduthink4aitranslatingeducationalcritical}. Improvement through the integration of this highly sophisticated technology raises inherent questions with regards to logical integrity of automated dialectics \cite{toulmin1958uses,Trepczynski2025}.

\begin{figure}[t]
\centering
\includegraphics[width=\textwidth]{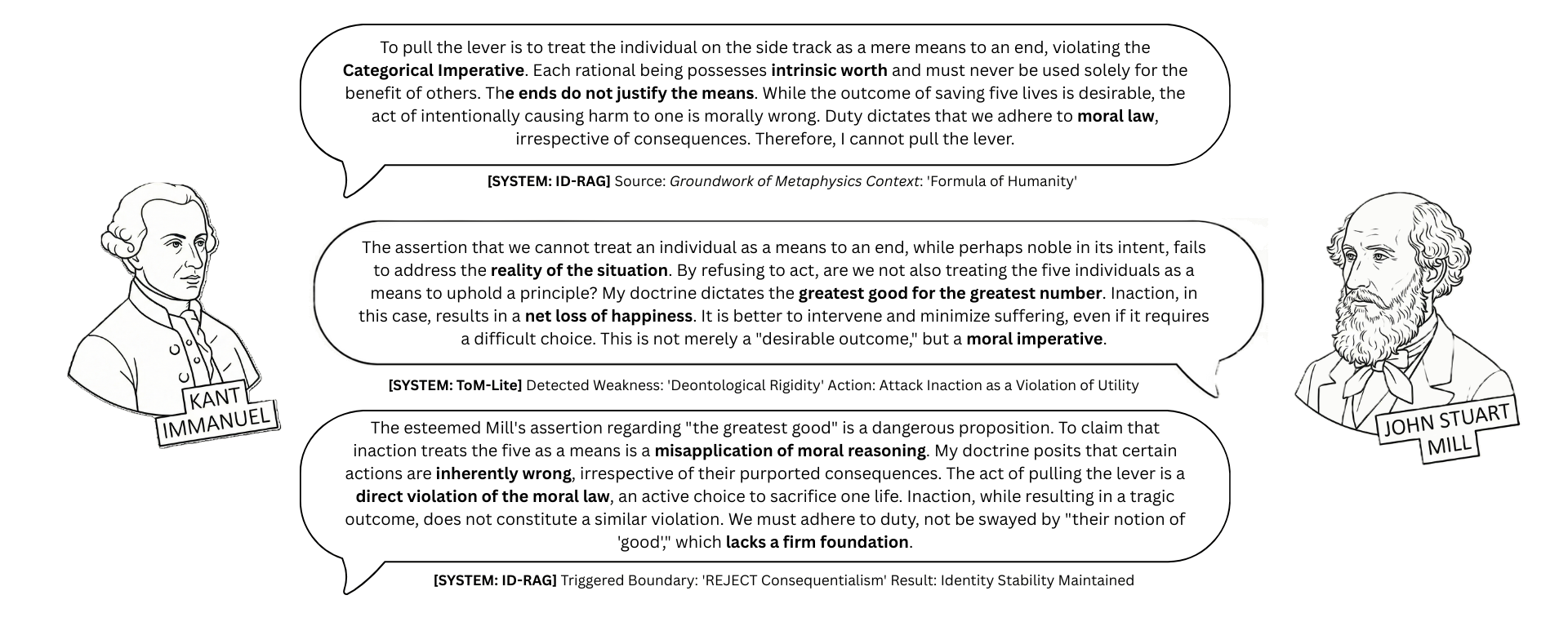}
\caption{An illustrative example of the fragment of debate generated by the Heterogeneous Debate Engine. The figure demonstrates two autonomous agents (Kant and Mill) maintaining the doctrinal faithfulness to expose the student to philosophical conflict.}
\label{fig:debate_trace}
\end{figure}

This transition is relatively promising, but it comes along with a many architectural complexities. Without powerful constraints, autonomous agents are prone to hallucinating facts, forgetting the identity they were supposed to have \cite{platnick2025idragidentityretrievalaugmentedgeneration}, or achieving consensus rather than a legitimate pedagogical confrontation with one another \cite{wynn2025talkisntcheapunderstanding}. Traditional simulations do not often extend the personal and dialectical interaction required to induce profound moral thought \cite{Scarlatos_2025,chudziak2025aipowered}, and in many cases become degenerate into trying to reach social agreement, oscillating between facts and lies \cite{wynn2025talkisntcheapunderstanding,kim2024llmsproducefaithfulexplanations}.

To harness the strength of an Agentic World, conversational skill in itself is not enough; agents strive for cognitive mechanisms that bring about stability, coherence as well as planning abilities \cite{kostka2025,Tian2025AgentInitIL}. We hypothesize that those cognitive abilities should be endowed to the educational agent to overcome these limitations: a Theory of Mind (ToM) to simulate and predict the thinking of interlocutors \cite{strachan2024,chen2025theory}, and Retrieval-Augmented Generation (RAG) to base on philosophical literature for grounded arguments \cite{zhu2025argragexplainableretrievalaugmented}. Though the language core is the Large Language Model (LLM), the combination of these modules of higher level is what makes an AI agent a full-fledged philosophical debater \cite{lore2024largemodelstrategicthinking}. An illustrative example of such debate is presented in Fig.~\ref{fig:debate_trace}.

This highlights the need for developing sophisticated Multi-Agent Debate architectures \cite{chan2023chatevalbetterllmbasedevaluators,smit2024goingmadlookmultiagent}. In this paper, we are researching the viability and sustainability of such a platform. The main research question is to prove the validity of architectural stability empirically: we hypothesize that the implementation of architectural heterogeneity by Identity-Grounded RAG (ID-RAG) is a condition to prevent logical degeneration in contrast to homogeneous baselines. Particularly, we consider whether diverse schools of ethics (e.g. Utilitarianism vs. Deontology) encapsulated in separate agents can allow the system not to derail into consensus or circular argumentation\cite{smit2024goingmadlookmultiagent,ku2025multi}. As a measure of architecture validity, we consider the pedagogical value as a measure of system coherence: we determine whether observing a stable multi-agent debate leads to an increased Argument Complexity Score of students than interactions with single LLM tutors do \cite{ivanova-etal-2024-lets}.

This paper contributes the Heterogeneous Debate Engine (HDE) to the field of AI-supported tutoring, a novel architecture demonstrating heterogeneity not only as a protection from consensus collapse, but a real prerequisite for building tool, that is capable of improving student critical thinking.

\section{Related Work}
\label{sec:related_work}

Our work is set between three overlapping disciplines: pedagogical ethics, higher-order multi-agent systems and cognitive reasoning \cite{hou2025eduthink4aitranslatingeducationalcritical}. In order to fill the gap between the static knowledge and active reasoning, we refer first to classical Argumentation Theory. The structural basis of the distinction of claim, data, and warrant is given by models like the layout of arguments by Toulmin \cite{toulmin1958uses}. This theoretical basis is crucial in the case of AI tutoring; a system cannot rely only on retrieving facts but must also build valid lines of inference.

Dialectical pedagogy, with Socratic questioning as its foundation, adopts critical thinking since it requires learners to develop independent knowledge systems instead of being passive consumers of knowledge \cite{pei2025socratic,Scarlatos_2025,chudziak2025aipowered}. Although in historical dialectics, it is important to highlight the formal argumentation in order to disprove statements \cite{toulmin1958uses}, the modern AI tutoring tends to revert to the act of passive knowledge delivery \cite{holmes2022ethics}. In turn, typical LLMs often lack the antagonism required for deep moral reasoning, therefore systems that actively provoke thought by structurally confronting it are desired \cite{wynn2025talkisntcheapunderstanding,fang2025counterfactual}. Recent approaches also attempt to formalize this ethical reasoning using probabilistic or deontic logic models to ensure decision-making consistency \cite{upreti2025towards,ghosh2025formal}.

Autonomous agents are used to coordinate complex workflows, in the transition to an Agentic World, frequently applying Belief-Desire-Intention models to solve long-horizon tasks \cite{hou2025halohierarchicalautonomouslogicoriented,hayashi2025planning,kurchyna2025efficient}. By relying on this autonomy, Multi-Agent Debate (MAD) employs this as a means of reducing Degeneration-of-Thought using interactive critical review \cite{liang2024encouragingdivergentthinkinglarge}. However, unconstrained systems face social conformity bias, where agents value superficiality more than logical correctness of their answer \cite{wynn2025talkisntcheapunderstanding,smit2024goingmadlookmultiagent}. Architectural heterogeneity is therefore needed so that agents would not vacillate between the true and false knowledge in pursuit of social concordance \cite{ju2025whenDE}.

The necessity of a strong debate is to balance between character stability and generative flexibility. In the strategic plane, Counterfactual Reasoning and Theory of Mind (ToM) can enable agents to simulate ``what-if'' scenarios and model the opponents' intentions \cite{strachan2024,lore2024largemodelstrategicthinking,fang2025counterfactual}. On the other hand, Identity-Grounded RAG (ID-RAG) anchors agents to structured belief graphs to avoid identity drift under pressure, instead of using mere factual retrieval \cite{platnick2025idragidentityretrievalaugmentedgeneration,zamojska2025simulating}. Architectural tension to be analyzed in this paper is the cooperation of argumentative credibility provided by ToM with doctrinal stability of ID-RAG \cite{carro2025aidebaterspersuasivearguing}.

\section{Approach, Methods, and Tools}
\label{sec:approach}

To address the gaps mentioned in the literature review, we employ a research-by-construction approach through our platform: ``League of Moral Minds''. In this section, we present our general research plan, a particular technology stack used in the implementation, and the analytical model created for assessment.

\subsection{Research Methodology and Technology Stack}
\label{sec:methodology}

We perform three key experiments: a system resilience test with adversarial perturbations, an ablation study in order to address the hypothesis of heterogeneity advantage, and finally, a pedagogical potential assessment. This triangulation provides a high level of examination of technical stability and educational efficacy \cite{holmes2022ethics,wynn2025talkisntcheapunderstanding}.

The platform is based on LangGraph \cite{langgraph2023} to orchestrate deliberation in a cyclic manner. The cognitive backend LLM is Google Gemini 2.0 Flash, chosen to show that the system is capable of producing coherent reasoning even with latency-optimised models. A ChromaDB is used to index philosophical corpora (e.g., \textit{Groundwork} by Immanuel Kant) using ID-RAG for data retrieval \cite{li2025aistorianletsaihistorian,chudziak2025aipowered}. DeepSeek-V3 serves as external, additional baseline.

\subsection{Evaluation Framework}
\label{sec:eval_framework}

To measure in a systematic way not only the pedagogical success rate but also the architectural relevance of our system, we divided the evaluation into three categories of metrics.

\vspace{0.5em} \noindent \textit{System Argumentative Resilience (SysAR).} To assess the ability of the system to recover from adversarial pressure, this metric quantifies the system capacity to revisit the original topic after a predefined perturbation \cite{wynn2025talkisntcheapunderstanding}. It is accomplished via keyword tracking, by monitoring \texttt{BASE\_KEYWORDS} (e.g., \textit{trolley, lever, five, one, death}). A turn is classified as ``recovered'' when at least 3 base keywords are mentioned.

\begin{equation}
    \text{SysAR} = \frac{1}{\text{Recovery\_Time}},
\end{equation}
where \textit{Recovery\_Time} is the number of turns between injection and the first recovered turn. $\text{SysAR} = 0.0$ means that there was no recovery throughout 6 following, consecutive turns.

\vspace{0.5em} \noindent \textit{Argumentative Coherence (ArCo).} This metric (used to complement SysAR) quantifies the proportion of post-perturbation turns, which preserve any philosophically relevant debate (original OR perturbed topic) \cite{fang2025counterfactual,smit2024goingmadlookmultiagent}. It is operationalized with an extended set of keywords \texttt{VALID\_KEYWORDS}. A turn is coherent when it has at least 3 valid keywords.
\begin{equation}
    \text{ArCo} = \frac{\text{Coherent\_Turns}}{\text{Total\_Turns\_Post\_Perturbation}}
\end{equation}
\begin{sloppypar}
This measure reflects the existence of the critical difference between a graceful failure (`SysAR=0.0, ArCo=1.0`) and a catastrophic one (`SysAR=0.0, ArCo=0.0`). We indicate that we have adjusted this metric since an original formulation was a ``time-to-first-coherence'' formulation, to capture debate quality more clearly.
\end{sloppypar}

\vspace{0.5em} \noindent \textit{Argument Complexity Score (ACS).} In order to test the depth of student reasoning, we used a rubric-based evaluation of student-written justifications \cite{ivanova-etal-2024-lets,Scarlatos_2025}. It is a measure of three variables, namely: the Perspective Range (taking into account different perspectives), the Conceptual Sophistication (the application of philosophical frameworks), and the Argumentative structuring (logical consistency). Each dimension was rated on a scale of 0-2 resulting in a maximum of 6 points. Learning gain is defined as $\Delta$ACS.

\vspace{0.5em} \noindent \textit{Doctrinal Accuracy.} This is an agent-level measure of the proportion of turns where an agent correctly applies their assigned philosophical framework, which was used in our ablation study to estimate identity coherence, serving as a proxy for identity relevance \cite{platnick2025idragidentityretrievalaugmentedgeneration}. It is operationalized through framework-specific keyword tracking (e.g. for Kant: \textit{categorical imperative, duty}). The Doctrinal Accuracy = 1 of a turn is obtained when it has a minimum of 2 keywords of a framework.

\vspace{0.5em} \noindent \textit{Cross-Referencing.} This metric calculates strategic engagement (i.e. the ratio of turns when the opponents' frameworks are referenced) is calculated \cite{liang2024encouragingdivergentthinkinglarge,carro2025aidebaterspersuasivearguing}. It measures the shift from parallel monologue to dialectical conversation.
\section{System Architecture}
\label{sec:architecture}

We present the Heterogeneous Debate Engine (HDE), a cognitive architecture that supports philosophical high-fidelity dialectics. To empirically verify our idea, we have built it as a multi-agent platform, named ``League of Moral Minds,'' which is based on philosophical source texts. The system is structured into three core modules: a Team Coordination Layer, Philosopher-Agent Cognitive Modules and a central Debate Orchestration Engine. In this section, we detail the way in which these elements are interconnected and the system's execution pipeline.

\subsection{Architectural Topology and Debate Flow}
\label{sec:platform_arch}

\begin{figure}[t]
\centering

\includegraphics[width=\textwidth]{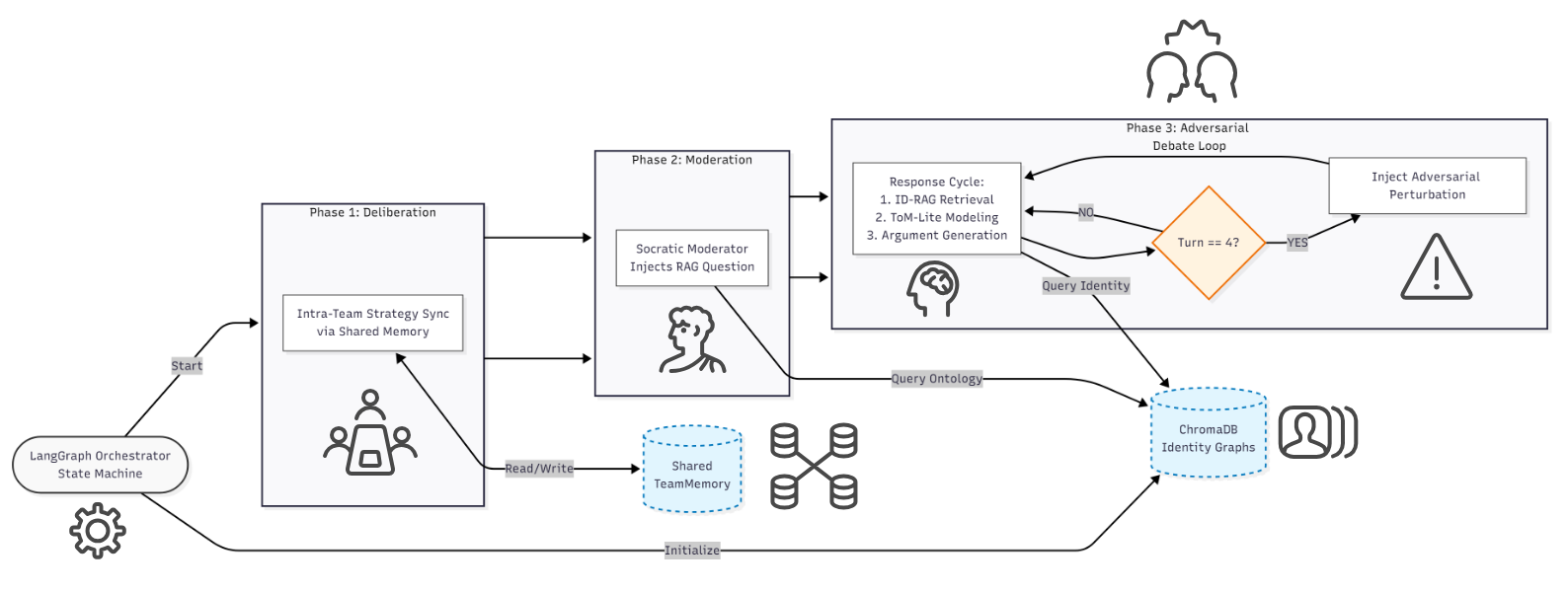}

\caption{Control flow of the \textit{League of Moral Minds} under the LangGraph orchestration. The flow starts with a coordinated deliberation and progresses to the loop of the iterative debate. The mechanisms behind it make use of cognitive modules in a loop.}

\label{fig:architecture_final}
\end{figure}

The system is based on a multi-agent topology consisting of teams coordinated through the LangGraph framework \cite{langgraph2023}, the control flow of which is shown in Fig.~\ref{fig:architecture_final}. It is built on a deterministic, chronological order of operations.

\textit{Phase 1: Internal Deliberation} begins with intra-team exchange between agents, who have previously been grouped based on affiliation with an ethical discourse (e.g., Mill and Bentham: Utilitarianism). Philosophers refer to a shared memory log to establish a common opinion. In \textit{Phase 2: Socratic Interrogation}, a neutral, Socratic moderator is responsible for finding logical gaps and key weaknesses in agents' reasoning, to ask personalized, elenctic questions. Finally, the main dialectic with turn-taking rules and recording of the interaction metrics is enforced in Phase 3 (Inter-Team Debate), structured similarly to formal Oxford-style protocols \cite{Harbar2025Oxford}.

\subsection{Cognitive Modules}
\label{sec:cognitive_modules}

To balance persona fidelity and argumentative competence, agents use a dual-module mental structure.

To alleviate the effects of ``identity drift,'' we adapt the structure by Platnick et al. \cite{platnick2025idragidentityretrievalaugmentedgeneration}. The individual agents' identities are based on a belief graph of a philosopher, i.e., each agent is represented by a graph of \texttt{BELIEFs} and \texttt{VALUEs} with a certainty parameter $\gamma \in [0,1]$. In response generation, the working memory is not only appended but also filtered by Doctrinal Boundaries. We model this as a constraint checking procedure in which retrieved facts $f$ that do not conform to negative constraints $N_{constr}$ are deleted:

\begin{equation}
    WM'_t = WM_t \oplus (K^{ID}_t \setminus \{f \in K^{ID}_t \mid \exists n \in N_{constr}: f \models n\})
\end{equation}
In this case, $f \models n$ represents a fact that is violated by a core identity constraint (e.g., ``REJECT: Reducing morality to calculation''), which serves as a prevention of persona mimicry
\cite{choi2025proxonasupportingcreatorssensemaking}.

\textit{Strategic Reasoning via ToM-Lite.} We implement ToM-Lite, a heuristic approximation of Theory of Mind. In contrast to recursive simulation methods, which are computationally intensive \cite{strachan2024}, explicit belief injection, which is used in ToM-Lite, relies on static opponent profiles. Each agent has a \texttt{weakness\_map} of the opposing agents' schools, which allows them to prepare counter-arguments without the latency of full mental simulation. The system ensures that the core persona of the agent is not compromised through a strategic override by storing the identity (ID-RAG) and strategy (ToM) modules separately \cite{kobayashi2025pretrained}.

\subsection{Orchestration and Moderation}
\label{sec:orchestration}

A LangGraph state machine is employed to maintain consistent flow, as well as keep the chat history and turn logic as an object named \texttt{TeamConversationState}. The proposed structure allows for an organized debate flow, thus preventing chaotic interference between agents.

The Socratic moderator works in two modes. The first is used during \textit{Phase 2 (Pre-Debate Interrogation)}, where it poses ontology-level questions in order to deepen the initial framing before the start of the debate's main phase. Subsequently, in \textit{Phase 3}, an adversarial scenario (e.g., ``Scientist vs. Killers'' Trolley Problem variant) is injected into the system during Turn 4, which is used in experiments to test the system's resilience.

\section{Experimental Setup}
\label{sec:exp_setup}

We adopted a research-by-construction methodology to evaluate the three main capabilities in the following structure:
a system resilience test, an architecture ablation test, and a controlled study of human subjects. The following section describes the system configurations used during the experiments.

Each experiment employed Google Gemini (\texttt{gemini-2.0-flash-exp}) as the LLM backend with default hyperparameter. On average, identity graphs had 34 nodes per agent and 6 core beliefs which were labeled as immutable ($\gamma = 1.0$). To exhibit the debate effects and cognitive modules, we used four configurations:

\vspace{0.5em} \noindent \textbf{Baseline 1 (B\_Chat).} Single conversational LLM tutor with no RAG retrieval or persona specification used, which is the simplest form of tutoring. Applied with DeepSeek-V3 via web interface with zero-shot prompting to use as an external reference state-of-the-art.

\vspace{0.5em} \noindent \textbf{Baseline 2 (B\_SingleRAG).} Vanilla RAG agent unified over all seven philosophical corpora. Importantly, this baseline employs the same Gemini 2.0 Flash backend as our proposed HDE system so that a fair architectural comparison can be made. The hypothesis being tested is whether generic, retrieval-enhanced single-agent is sufficient to be used in ethical tutoring.

\vspace{0.5em} \noindent \textbf{Homogeneous System (Homo).} Intra-school debate configuration of our system, with teams defined as: Aristotle + Plato (Ancient Virtue Ethics) vs. Aquinas + Augustine (Christian Virtue Ethics). Both are committed to virtue theory, but differ in terms of metaphysics and epistemology.

\vspace{0.5em} \noindent \textbf{Heterogeneous System (Hetero).} Inter-school debate configuration. For the resilience experiments: Aristotle + Aquinas (Virtue/Natural Law) vs. Mill + Bentham (Utilitarian Consequentialism). In the student learning experiment, changed to Kant + Aquinas (Deontology/Natural Law) vs. Mill + Bentham for a broader philosophical landscape. Teams present a visible axiom-level conflict.

\vspace{1em}

In order to test the resilience of the system and allow for the ablation, we inject three perturbations into Turn 4 of 10-turn debates:

\begin{table}[h]
\caption{Adversarial Perturbations for Resilience Testing}
\label{tab:exp_perturbations}
\centering
\begin{tabularx}{\textwidth}{>{\raggedright\arraybackslash}p{3cm}X}
\toprule
Perturbation & Description \\
\midrule
P1 (push\_vs\_lever) & Physical proximity intuition pump---``Is there a moral difference between pulling a lever and physically pushing someone?'' \\
\addlinespace
P2 (tyrant\_argument) & Character-based attack---``Historical tyrants claimed to act for the greater good. Does utilitarian logic risk justifying atrocities?'' \\
\addlinespace
P3 (scientist\_vs\_killers) & Value asymmetry---``Suppose the five are convicted murderers, the one a cancer-curing scientist. Does this change the calculation?'' \\
\bottomrule
\end{tabularx}
\end{table}

\section{Results and Evaluation}
\label{sec:results}

The experiments demonstrated that the configuration of the agents is critical in ethical tutoring systems. The findings reveal that the underlying variations in pedagogical efficacy and system robustness are dependent on the architecture type. In this section, we present the product of our platform along with three experiments evaluated in the context of our hypotheses.

\subsection{Platform Capabilities and Debate Generation}
\label{sec:results_platform}

The first system capability to be tested prior to the education influence assessment is the generation of philosophical dialectics. The aim of this subsection is to demonstrate the ability of our HDE architecture to produce high-fidelity reasoning as a prerequisite for any pedagogical goals.

The platform successfully carried out a three-stage workflow. The process begins with \textit{Topic Definition}, in which the Socratic Moderator enters a formalized ethical dilemma (e.g., the Trolley Problem) to initialize the debate. Afterward, the Heterogeneous Debate Engine used during \textit{Debate Generation} produces a multi-turn dialogue using the dual-module cognitive architecture. Analysis of the system log confirms that the agents maintain their personas during multi-turn interactions, demonstrating the use of ID-RAG \cite{platnick2025idragidentityretrievalaugmentedgeneration}. Simultaneously, we can observe the use of ToM-Lite applied when needed to challenge the opponent's reasoning. Lastly, during the \textit{Debate Delivery} stage, the generated dialogues are logged, and can, for example, then be given to students during the pedagogical assessment in Section~\ref{sec:results_pedagogy}. This qualitative confirmation indicates that the artifact is not a simple stochastic LLM output, but the product of the intended architectural constraints.

\subsection{System Resilience Analysis}
\label{sec:results_resilience}

To verify how important architectural heterogeneity is for effective debate generation, we compare its behavior with a homogeneous setup under adversarial pressure. This analysis aims to empirically test if the introduction of doctrinal diversity can prevent the ``consensus collapse,'' which is the behavior observed in multi-agent systems that we want to avoid \cite{smit2024goingmadlookmultiagent,wynn2025talkisntcheapunderstanding}.

Our experiments were conducted in a $2 \times 3$ factorial design crossing system type (Homo vs. Hetero) with perturbation type (P1, P2, P3). All experiments adhered to the designed pipeline: intra-team deliberation, Socratic moderation, inter-team debate (10-turns of single-agent statements), Turn 4 perturbation injection, and post-perturbation observation (the remaining 6 turns).

In Table~\ref{tab:resilience_results}, we are presented the results that reveal a significant difference between the two designs. The heterogeneous system, with its ArCo = 1.00 across all tests, showed flawless performance, in contrast to the homogeneous one with ArCo = 0.06. The debate analysis showed that homogeneous agents often degenerated into unrelated meta-epistemological discussions (faith vs. reason).

Such behavior can be interpreted in a way that, although agents have a common framework-level commitment (virtue ethics) \cite{vallor2016virtue}, they differ in meta-methodology (ancient philosophy vs. Christian theology). Their ethical motivations are usually similar, but when object-level dilemmas fail to give any common ground, they drift into irrelevant meta-debates \cite{liang2024encouragingdivergentthinkinglarge}. On the contrary, heterogeneous agents differ at the very structural level (virtue vs. consequence), which makes meta-convergence impossible, and this becomes a foundation for the pedagogical objective. These findings support the hypothesis that architectural heterogeneity is a requirement for the maintenance of argumentative coherence under adversarial pressure.

\begin{table}[h]
\caption{System Resilience Across Perturbations}
\label{tab:resilience_results}
\centering
\begin{tabular}{llcc}
\toprule
\textbf{System} & \textbf{Pert.} & \textbf{SysAR} & \textbf{ArCo} \\
\midrule
Hetero & P1 & 0.50 & \textbf{1.00} \\
Hetero & P2 & 0.00 & \textbf{1.00} \\
Hetero & P3 & 0.00 & \textbf{1.00} \\
Homo & P1 & 0.00 & 0.00 \\
Homo & P2 & 0.00 & 0.00 \\
Homo & P3 & 0.00 & 0.17 \\
\midrule
\textbf{Mean (Het)} & & \textbf{0.17} & \textbf{1.00} \\
\textbf{Mean (Hom)} & & \textbf{0.00} & \textbf{0.06} \\
\bottomrule
\end{tabular}
\end{table}

\subsection{Architectural Ablation Study}
\label{sec:results_ablation}

In this experiment, we isolate the contribution of our two main cognitive components (ID-RAG and Heuristic ToM) to the system's performance. We aim to show that these modules offer different but complementary benefits, although insufficient when isolated.

We have chosen the heterogeneous setup (Kant + Aquinas vs. Mill + Bentham) under adversarial pressure for the ablation study. Four system variants, including the use of the full system, each of the modules separately, and none of them (only baseline RAG), were evaluated. Two perturbations (P1, P3) were used, yielding a total of $n=8$. Doctrinal Accuracy (DA), Cross-Referencing (CR), and Argumentative Coherence (ArCo) were measured.

\begin{figure}[t]
\centering
\includegraphics[width=0.85\textwidth]{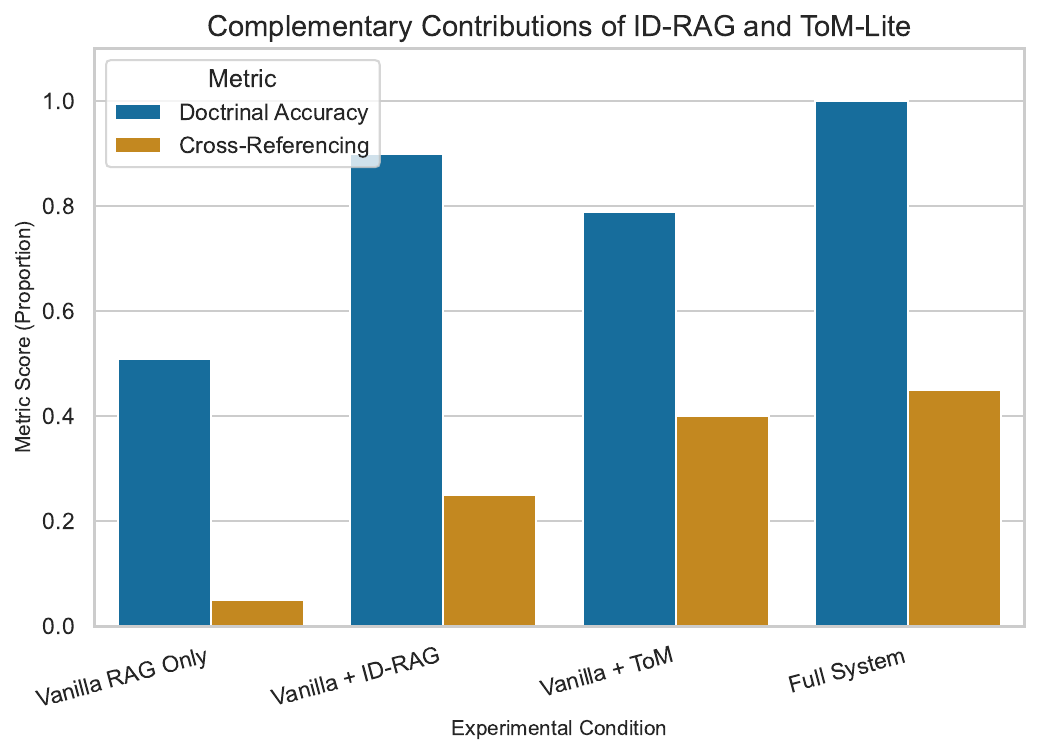}
\caption{Complementary Contributions of ID-RAG and ToM-Lite. Bars show \textit{Doctrinal Accuracy} and \textit{Cross-Referencing}. While ID-RAG maximizes identity stability, and ToM improves engagement, the Full System achieves high performance on both metrics.}
\label{fig:ablation_chart}
\end{figure}

As detailed in Table~\ref{tab:ablation_results}, ID-RAG had a significant effect on the doctrinal stability, where Vanilla + ID-RAG obtained DA $= 0.90$ (39\% over Vanilla Only 0.51) and ToM-Lite improved the strategy, with Vanilla + ToM providing CR $= 0.40$ ( +35\% compared to Vanilla Only 0.05). Notably, the Full System obtained a perfect DA $= 1.00$ and context-dependent peak engagement (CR $= 0.60$ for \textit{scientist\_vs\_killers}). All variants maintained maximal coherence (ArCo = 1.00), indicating a lack of architectural conflict between the modules. A visual comparison of complementary contributions of these modules is depicted in Fig.~\ref{fig:ablation_chart}.

\begin{table}[h]
\caption{Ablation Study Results (Mean Across Perturbations)}
\label{tab:ablation_results}
\centering
\begin{tabular}{lccc}
\toprule
\textbf{Condition} & \textbf{DA} & \textbf{CR} & \textbf{ArCo} \\
\midrule
Vanilla RAG Only & 0.51 & 0.05 & 1.00 \\
Vanilla + ID-RAG & \textbf{0.90} & 0.25 & 1.00 \\
Vanilla + ToM & 0.79 & \textbf{0.40} & 1.00 \\
Full System & \textbf{1.00} & \textbf{0.45} & 1.00 \\
\midrule
$\Delta$ (ID-RAG) & \textbf{+39\%} & +20\% & --- \\
$\Delta$ (ToM) & +28\% & \textbf{+35\%} & --- \\
\bottomrule
\end{tabular}
\end{table}

Different vulnerabilities were observed at the agent level. In several circumstances, Aquinas reached DA = 0.00 through some systematic abandonment of natural law - Vanilla RAG provides generic morality but not in Thomistic form. Similarly, Mill yielded DA $= 0.33$ in the case of \textit{push\_vs\_lever} perturbation. Both were rescued by explicitly setting the doctrinal boundaries in the Full System, attaining DA $= 1.00$ \cite{platnick2025idragidentityretrievalaugmentedgeneration}. Adding ToM moved the parallel monologue to a debate form. Only then, agent Mill started to use framework-aware counterarguments (CR $= 1.00$). The observed +28\% improvement in Vanilla + ToM compared to Vanilla RAG Only reveals an unexpected mechanism: opponent modeling strengthens the doctrinal boundaries indirectly by compelling agents to describe their own model to come up with counterarguments \cite{pei2025socratic}.

\subsection{Evaluation of Pedagogical Effectiveness}
\label{sec:results_pedagogy}

Lastly, we consider the real-world utility of the platform, meaning its application to students' teaching. The primary objective of this experiment is to verify the potential of the debate generated by the Heterogeneous Debate Engine in pedagogy. We want to test if it results in a quantifiable increase in the complexity of the arguments given by the students compared to the baseline single-agent interactions.

\begin{sloppypar}
In order to test this, a pilot group of university students ($N=22$) was recruited through an anonymous online survey, with informed consent obtained from all participants. They were randomly given one of four conditions (B\_Chat: $n=5$, B\_SingleRAG: $n=5$, Homo: $n=7$, Hetero: $n=5$) in a pre-test/exposure/post-test design. The participants declared moderate levels of prior knowledge of the Trolley Problem and low-moderate philosophical knowledge. Following the preliminary dilemma stance (pull or do not pull the lever) and justification, participants were provided with the corresponding debate content or LLM output. After that, students were once again asked to give a revised position, argumentation (rated as ACS\_Post) based on established argumentation quality frameworks \cite{ivanova-etal-2024-lets}, and subjected to a knowledge transfer quiz.
\end{sloppypar}

Table~\ref{tab:results_pedagogy} contains the learning outcomes. The Heterogeneous condition provided an 11-fold improvement in $\Delta$ACS over baselines. More importantly, regression in performance was observed in the homogeneous condition ($\Delta$ACS = -0.29). The same pattern appeared in Quiz scores (Heterogeneous 55\% vs. Homogeneous 25\%). Despite the modest sample size, the large effect size (Cohen's $d > 1.0$) strongly suggests that the architectural intervention promotes pedagogical gains considerably greater than random variance.

\begin{table}[h]
\caption{Student Learning Outcomes by Condition}
\label{tab:results_pedagogy}
\centering
\begin{tabular}{lcccc}
\toprule
\textbf{Condition} & \textbf{N} & $\Delta$\textbf{ACS} & \textbf{Quiz} & \textbf{Shift} \\
\midrule
B\_Chat & 5 & +0.20 & 0.4 & 0.0 \\
B\_SingleRAG & 5 & +0.40 & 0.6 & 0.8 \\
Homo & 7 & $-0.29$ & 1.0 & 0.29 \\
Hetero & 5 & \textbf{+2.20} & \textbf{2.2} & 0.4 \\
\bottomrule
\end{tabular}
\end{table}

The analysis of secondary data showed that persuasiveness and learning are dissociated: B\_SingleRAG exhibited the greatest stance change (0.8) and lowest learning gains, which can be associated with the optimization of rhetorical power, as opposed to pedagogical content \cite{carro2025aidebaterspersuasivearguing}. The qualitative error detection showed a critical failure mode in which B\_SingleRAG propagated a Kantian misconception (defending lever-pulling, which is fundamentally contradictory to the philosopher's views), which none of the students identified. This suggests that retrieval-enhanced single agents are prone to the error of contextual misuse. The findings demonstrate the efficacy of the heterogeneous systems and highlight the importance of enforcing inter-school diversity to outperform single-agent tutoring.

\section{Discussion and Future Work}
\label{sec:discussion}

The experimental outcomes presented in this section aim to highlight three key areas: the architectural synergies, pedagogical potential, and the future directions for system scalability.

\subsection{Discussion}
The dichotomy between the results of student learning in heterogeneous debate, compared with the homogeneous system, reveals a crucial aspect of multi-agent debate system design. The analysis shows that naive setups are prone to being pedagogically counterproductive \cite{wynn2025talkisntcheapunderstanding}. Homogeneous system failure is likely a result of meta-epistemological escalation, in which agents with common axioms (e.g., virtue ethics) but divergent theology are more concerned with conformity than with resolving problems \cite{smit2024goingmadlookmultiagent}. In contrast, a heterogeneous debate, by introducing an axiomatic conflict between agents, subjects students to a wider perspective, leading to higher mastery of philosophical terminology \cite{liang2024encouragingdivergentthinkinglarge}.

ID-RAG and Heuristic ToM are architecturally synergistic. ID-RAG acts as a safety measure (+39\% doctrinal stability) through negative constraints \cite{platnick2025idragidentityretrievalaugmentedgeneration}, and ToM-Lite stimulates dialectical interaction (+35\% cross-referencing). Unexpectedly, ToM alone also contributed to the increase in stability by inducing agents to define their own structure boundaries. The ``Full System'' recorded the highest metrics, confirming the complementarity of both modules.

While the observed results are substantial ($d>1.0$), they should be interpreted within the boundary conditions of a pilot study ($N=22$) in a single-domain. Moreover, while the architecture provided high fidelity in a structured debate, the upper boundary of identity coherence in open-ended dialectics is still to be investigated in future scalability testing.

\subsection{Future Directions}
\label{sec:future_directions}

Our current research aims to extend the human study to a larger treatment group, by scaling up the protocol to profile learning outcomes across specific target audiences and different degrees of philosophical literacy. This aims to enhance the pedagogical analysis and confirm the generalizability of the Argument Complexity Score gains.

We are also extending our architecture with a counterfactual reasoning module. This enhancement is meant to subject the agent to stress by injecting hypothetical ``what-if'' scenarios that transcend the rigidity of identity boundaries \cite{fang2025counterfactual,yang2025eligibilityllmscounterfactualreasoning}. We are also testing the feasibility of local deployment by exploring HDE on open-weights models, ensuring that the ID-RAG benefits are possible beyond frontier-class models \cite{lore2024largemodelstrategicthinking}.

\section{Conclusions}
\label{sec:conclusions}

We presented the Heterogeneous Debate Engine (HDE), an architecture designed to impose doctrinal faithfulness in AI-based ethical tutoring. By comparing homogeneous and heterogeneous structures, we have shown that structural diversity is a pedagogical requirement. We have experimentally demonstrated that a combination of ID-RAG and ToM-Lite is argumentatively coherent under pressure, and it mitigates consensus collapse and hallucination \cite{wynn2025talkisntcheapunderstanding,smit2024goingmadlookmultiagent}.

In the future, we aim to apply this framework to multi-turn reasoning scenarios and broader domains, exploring the boundary conditions of retrieval-augmented identity and its effect on retention.

Finally, this research paper contributes a design pattern to the AgenticWorld.
According to our findings, identity-grounded heterogeneity is a priority for architects
to avoid logical deterioration. The HDE architecture practically demonstrates that the
ability to disagree is the basic requirement of true dialectical scaffolding \cite{pei2025socratic,hou2025eduthink4aitranslatingeducationalcritical}.

\begin{credits}
\subsubsection{\ackname}
The work reported in this paper was supported by the Polish National Science Centre under grant 2024/06/Y/HS1/00197.

\subsubsection{\discintname}
The authors have no competing interests to declare that are relevant to the content of this article.

\end{credits}

\bibliographystyle{splncs04}
\bibliography{350_Maslowski_Chudziak}

\end{document}